\documentclass{article}
\usepackage{spconf,amsmath,graphicx}
\usepackage{multirow}
\usepackage{makecell}
\usepackage{arydshln}
\usepackage{booktabs}
\usepackage{embrac}
\usepackage{amssymb}
\usepackage{amsmath}
\usepackage{graphicx}
\usepackage{color}
\usepackage{bibspacing}
\usepackage{enumitem}
\usepackage{threeparttable}
\usepackage{diagbox}
\usepackage{setspace}
\usepackage{array,caption}
\usepackage[labelfont=bf]{caption}
\usepackage{hyperref}
\usepackage{ulem}
\title{A DYNAMIC GRAPH INTERACTIVE FRAMEWORK WITH LABEL-SEMANTIC INJECTION FOR SPOKEN LANGUAGE UNDERSTANDING}

\name{Zhihong Zhu,Weiyuan Xu, Xuxin Cheng, Tengtao Song, Yuexian Zou$^{*}$
\thanks{
*Corresponding author: zouyx@pku.edu.cn
}
}
\address{
   ADSPLAB, School of ECE, Peking University, Shenzhen, China
}
\begin{document}
\topmargin=0mm
%\ninept
%
\maketitle

\begin{abstract}
\textit{Multi-intent detection} and \textit{slot filling} joint models are gaining increasing traction since they are closer to complicated real-world scenarios. However, existing approaches (1) focus on identifying \textit{implicit} correlations between utterances and \textit{one-hot} encoded labels in both tasks while ignoring \textit{explicit} label characteristics; (2) directly incorporate multi-intent information for each token, which could lead to incorrect slot prediction due to the introduction of irrelevant intent. In this paper, we propose a framework termed \texttt{DGIF}, which first leverages the semantic information of labels to give the model additional signals and enriched priors. Then, a multi-grain interactive graph is constructed to model correlations between intents and slots. Specifically, we propose a novel approach to construct the interactive graph based on the injection of label semantics, which can automatically update the graph to better alleviate error propagation. Experimental results show that our framework significantly outperforms existing approaches, obtaining a relative improvement of 13.7\% over the previous best model on the MixATIS dataset in overall accuracy.

% Experimental results on two public multi-intent datasets demonstrate that our approach achieves the state-of-the-art performance.
\end{abstract}
\vspace{-0.25em}
\vspace{-0.25em}
\section{Introduction}
\begin{figure}[htb]
\begin{minipage}[a]{1.0\linewidth}
  \centering
  \centerline{\includegraphics[width=8.8cm]{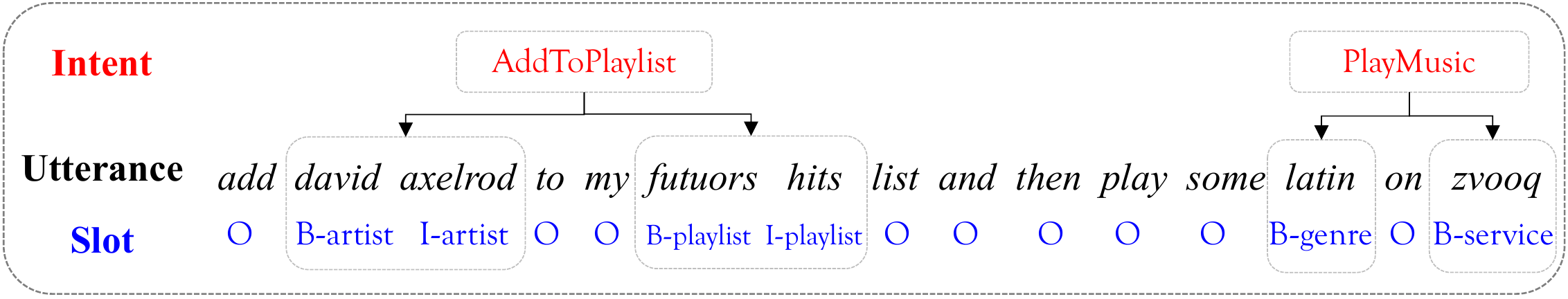}}
  \smallskip
\end{minipage}

\caption{Utterance with multiple intents (\textcolor{red}{red}) and slots (\textcolor{blue}{blue}).}
\label{fig:example}
\vspace{-1.5em}
\end{figure}

\vspace{-0.5em}
\label{sec:intro}
\textit{Spoken language understanding} (SLU) is a crucial component in task-oriented dialogue systems \cite{weld2021survey}, which typically consists of two subtasks: \textit{intent detection} (ID) and \textit{slot filling} (SF). \cite{kim2017fourth} discovered that complicated real-world scenarios frequently involve \textit{multiple} intents in a single utterance. Take an example in Fig.\ref{fig:example}, the task of ID should classify both intent labels in the utterance (\textit{i.e.}, \texttt{AddToPlaylist} and \texttt{PlayMusic}), while SF can be treated as a sequence labeling task to predict slot for each token in \texttt{BIO} format \cite{zhang2016joint,qin2021co}. 

% where utterance-level correlation is established implicitly.
Since intents and slots are inextricably related \cite{zhou2021pin,huang-etal-2020-federated, huang2021sentiment}, researchers in recent years \cite{gangadharaiah2019joint, qin-etal-2020-agif, qin2021gl,cai2022slim} have increasingly focused on joint \textit{multiple intent detection} and \textit{slot filling}. Although achieving promising performance, existing approaches typically classify an utterance to intents represented by \textit{one-hot} encoding(\textit{e.g.}, \texttt{0}) while the same problem occurs in \textit{slot filling}.
They ignore \textbf{intuitive} and \textbf{explicit} label characteristics, oversimplifying representations of labels. We argue that the label semantics may be useful, which could improve performances for both subtasks by assessing semantic similarity between words in utterances and words in labels.

Another key challenge in multi-intent SLU is how to effectively incorporate multiple intents information to \textit{guide} the slot prediction. To handle this, \cite{gangadharaiah2019joint} first investigated a multi-task network with a slot-gated mechanism \cite{goo2018slot}. For fine-grained interaction between multiple intents and slots, \cite{qin-etal-2020-agif} proposed an adaptive graph interactive framework, which builds an interactive graph for each token in the utterance by using all predicted intents. \cite{qin2021gl} explored a global-locally graph interaction network, which models slot dependency and intent-slot interaction for each utterance. However, different tokens appearing in the utterance have various importance for representing the intents. Unfortunately, models mentioned above straightforwardly attach multiple intent information to all tokens, including those without contribution to intent representations, which will introduce \textbf{noise} into sentence-level semantics to some extent.
% However, different tokens appearing in the utterance contribute variously to representing the intents.
% For example, in the utterance “\textit{what will be the wind speed in south sudan}”, the location of “\textit{south sudan}” is irrelevant to the judgment of intent “\textit{GetWeather}”. 
\begin{figure*}[t]
% \vspace{-0.5em}
  \centering
  \includegraphics[width=\linewidth]{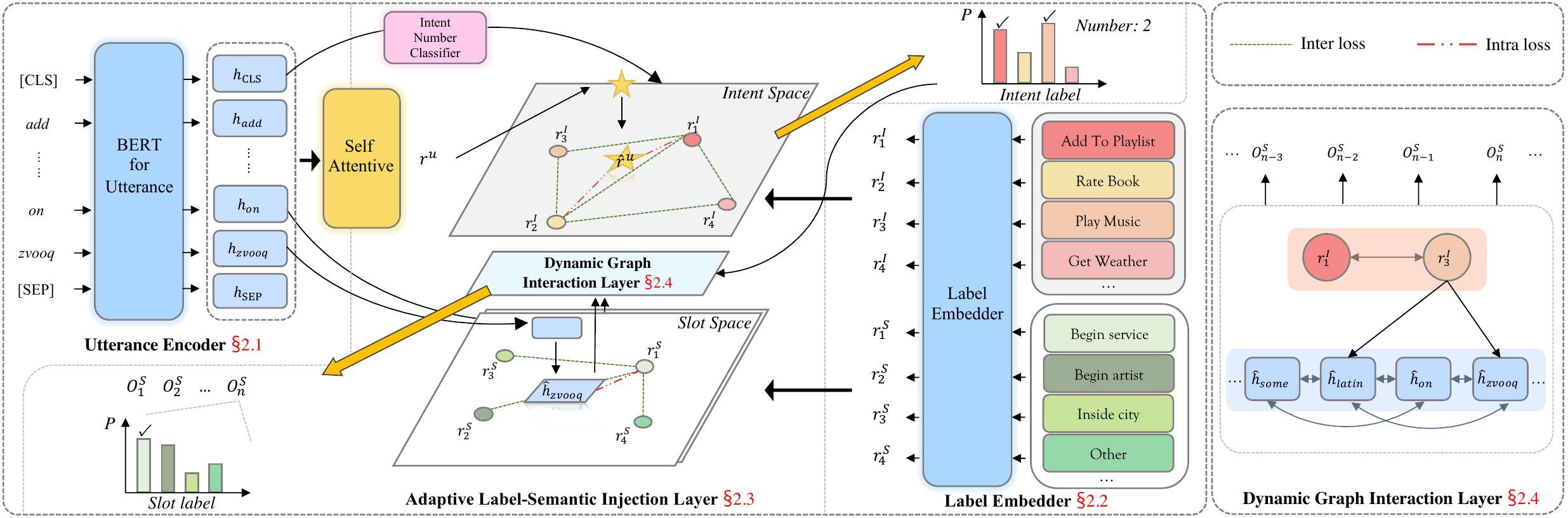}
  \caption{The architecture of \texttt{DGIF}. Two BERT encoders will encode both the \textit{utterance} (\textcolor{red}{\S\ref{sec:2.1}}) and \texttt{labels} (\textcolor{red}{\S\ref{sec:2.2}}). \textit{Sentence-level} representation 
  $r^u$ and \textit{token-level} representation $h_*$ will be injected with \texttt{label} semantics in respective space (\textcolor{red}{\S\ref{sec:2.3}}) and interact in \textcolor{red}{\S\ref{sec:2.4}}.
   $\checkmark$ denotes the \texttt{label} is selected as prediction. For simplicity, we only draw one case with \textit{several} \texttt{labels}.}
  \label{fig:network}
\vspace{-1em}
\end{figure*}
% Different color denotes different representation against \textit{utterance} $r^u$, \texttt{intents} $r^{I}_{*}$ and \texttt{slots} $r^{S}_{*}$.
%  In addition, we propose label-aware regularization to improve the relative distribution of labels representation, which construct correlation between labels.
% To overcome aforementioned restrictions

In this paper, we propose a novel framework \texttt{DGIF} for joint multiple ID and SF to tackle the above two issues. Concretely, inspired by the success of leveraging label characteristics to help model optimization \cite{wu-etal-2021-label,cui2019hierarchically}, we construct intent and slot spaces using words in each intent label and slot label respectively to inject label information into utterance representations adaptively. Moreover, we propose label-aware regularization to model the rich semantic dependencies among labels in each label space. Then, we capture relevant intents for each token to construct the multi-grain intent-slot interactive graph, as opposed to prior works which directly incorporate multiple intents information statically. Empirical results on two public datasets (MixATIS and MixSNIPS \cite{qin-etal-2020-agif}) demonstrate that our framework outperforms competitive baselines. The source code for this paper can be obtained from \href{https://github.com/Zhihong-Zhu/DGIF}{\texttt{https://github.com/Zhihong-Zhu/DGIF}}.
% Furthermore, our sentence-level graph can achieve 2 to 4 speed gain compared with its token-level counterpart \cite{qin-etal-2020-agif}.

\section{Methodology}
\vspace{-0.5em}
As shown in Fig.\ref{fig:network}, our framework consists of four major components, and we use a joint training scheme to optimize \textit{multiple intent detection} and \textit{slot filling} simultaneously. 

% Next, we will describe each component in detail.

% introduction
\label{sec:pro}
\vspace{-0.75em}
\subsection{Utterance Encoder}
\label{sec:2.1}
\vspace{-0.75em}
For the input utterance $\boldsymbol{U}=(u_1,...,u_n)$, we first prepended \texttt{[CLS]} and appended \texttt{[SEP]}, in order to match the input of BERT \cite{devlin2018bert}. Then, we employ a self-attentive network \cite{lin2017structured} over the output $\boldsymbol {h}=(\boldsymbol{h}_{\texttt{CLS}},\boldsymbol{h}_1,...,\boldsymbol{h}_n, \boldsymbol{h}_{\texttt{SEP}})$ of BERT's encoder to capture the sentence representation $\boldsymbol{r}^u$ with context-aware features, where $\boldsymbol{h}\in \mathbb{R}^{ (n+2) \times d}$ and $\boldsymbol{r^u} \in \mathbb{R}^d$.
\vspace{-0.75em}
\subsection{Label Embedder}
\label{sec:2.2}
\vspace{-0.75em}
We apply two steps to get label representations. Concretely, considering the ambiguity of label semantics, we first manually convert slot label names to their natural language forms (\textit{e.g.}, “\texttt{B-PER}” to “\texttt{begin person}”) while maintaining intent label names. Then, another BERT with self-attentive layer (\S\ref{sec:2.1}) is adopted to obtain label representation $\boldsymbol{r}_{i}^{\varphi}$, where $\varphi \in \{I,S\}$ ($I$ denotes intent label and $S$ denotes slot label). The reason to use a different BERT is that the utterance and labels commonly differ in syntactic structure.
\vspace{-0.75em}
\subsection{Adaptive Label-Semantic Injection Layer}
\label{sec:2.3}
\vspace{-0.75em}
\textbf{Label-Semantic Injection}\,
 We leverage \textit{best approximation} \cite{del1995statistical} idea to help incorporate label information into utterance representations. The \textit{best approximation} problem specifies that $\mathcal{T}$ is a subspace of Hilbert space $\mathcal{S}$. For a given vector $\boldsymbol{x} \in \mathcal{S}$, we need to find the closest point $\hat{\boldsymbol{x}} \in \mathcal{T}$. It turns out solution of $\hat{\boldsymbol{x}} =  {\textstyle \sum_{n=1}^{N}}a_n\boldsymbol{v}_n$ will be a linear combination of a basis $\boldsymbol{v}_1,...,\boldsymbol{v}_{N}$ for $\mathcal{T}$ of $N$ dimension. Coefficients $\boldsymbol{a}$ satisfies $\boldsymbol{Ga}=\boldsymbol{b}$ where $\boldsymbol{b}_n=\left \langle \boldsymbol{x},\boldsymbol{v}_n \right \rangle $ and $\boldsymbol{G}_{k,n}=\left \langle \boldsymbol{v}_n,\boldsymbol{v}_k \right \rangle $.
 
Instead of directly utilizing the $\boldsymbol{r}^u$ and $\boldsymbol{h}$ to predict the intent and slot labels, we first construct the label space $\mathcal{T}^{\varphi}$ with a basis $\{ \boldsymbol{r}_{1}^{\varphi},...,\boldsymbol{r}_{\left | \varphi  \right | }^{\varphi} \}$ by $\left | \varphi  \right |$ label embeddings in \S\ref{sec:2.2}. Then for a given $\boldsymbol{r}^{u^\varphi }$, we can project it onto $\mathcal{T}^{\varphi}$ to obtain its best approximation $\hat{\boldsymbol{r}}^{u^\varphi }= \boldsymbol{w}^{\varphi^\top} \boldsymbol{r}_{}^{\varphi}$, where
$\boldsymbol{w}^ \varphi=\boldsymbol{G}^{
\varphi^{-1}}\boldsymbol{b}^ \varphi$. The Gram matrix $\boldsymbol{G}^ \varphi$ and $\boldsymbol{b }^ \varphi$ are calculated as follows:
\begin{equation}
\setlength{\abovedisplayskip}{3pt}
\setlength{\belowdisplayskip}{3pt}
\scalebox{.84}{$
\boldsymbol{G}^\varphi = \begin{bmatrix}
  \left \langle \boldsymbol{r}_{1}^{\varphi},\boldsymbol{r}_{1}^{\varphi} \right \rangle & \cdots  &\left \langle \boldsymbol{r}_{\left | \varphi  \right | }^{\varphi},\boldsymbol{r}_{1}^{\varphi} \right \rangle  \\
  \vdots & \ddots  & \vdots  \\
   \left \langle \boldsymbol{r}_{1}^{\varphi},\boldsymbol{r}_{\left | \varphi  \right | }^{\varphi} \right \rangle & \cdots  & \left \langle \boldsymbol{r}_{\left | \varphi  \right | }^{\varphi},\boldsymbol{r}_{\left | \varphi  \right |}^{\varphi} \right \rangle
\end{bmatrix} \quad
\boldsymbol{b}^\varphi = \begin{bmatrix}
 \left \langle \boldsymbol{r}^{u^\varphi }, \boldsymbol{r}_{1}^{\varphi} \right \rangle  \\
 \vdots\\ 
\left \langle \boldsymbol{r}^{u^\varphi }, \boldsymbol{r}_{\left | \varphi  \right | }^{\varphi} \right \rangle
\end{bmatrix}
$}
\end{equation}
in which $\boldsymbol{r}^{u^I}$
denotes $\boldsymbol{r}^u$ and $\boldsymbol{r}^{u^S}$ denotes each token of $\boldsymbol{h}$.
% \begin{equation}
% \setlength{\abovedisplayskip}{3pt}
% \setlength{\belowdisplayskip}{3pt}

% \end{equation}

\noindent\textbf{Label-Aware Regularization}\,
While the label-semantic injection dynamically injects label information into utterance representations, we argue that it ignores the semantic dependencies among labels. Thus, we propose \textit{label-aware regularization} to ensure that the projected representation $\hat{\boldsymbol{r}}^{u^\varphi }$ capture the topological structure of the label space $\mathcal{T}^{\varphi}$ well. Concretely, we use \textit{Euclidean} and \textit{Cosine} distance to measure similarities, and optimize label representations as follows:
\begin{equation}
\setlength{\abovedisplayskip}{3pt}
\setlength{\belowdisplayskip}{4pt}
\scalebox{.76}{$
\begin{aligned}
\mathcal{L}_{inter}^{\varphi}=1+\frac{1}{Q\left | \varphi  \right | }\sum_{i=1}^{Q} \sum_{j=1}^{\left | \varphi  \right |}\operatorname{cos}(\boldsymbol{r}_{i}^{\varphi},\boldsymbol{r}_{j}^{\varphi})
\quad
\mathcal{L}_{intra}^{\varphi}=\frac{1}{PQ} \sum_{i=1}^{P} \sum_{j=1}^{Q}\left | \left | \boldsymbol{r}_i^{{u^\varphi }}-\boldsymbol{r}_{j}^{\varphi} \right |  \right | _{2}^{2}
\end{aligned}
$}
\end{equation}
where $P$ is the number of samples (or $n$ of $\boldsymbol{U}$), $Q$ is the number of gold labels in $\boldsymbol{r}_*^{{u^\varphi }}$. The regularization loss of $\mathcal{T}^{\varphi}$ is $\mathcal{L}_{{\rm RE}}^{\varphi}=\mathcal{L}_{inter}^{\varphi} + \lambda   \times \mathcal{L}_{intra}^{\varphi}$ and $\lambda $ is a hyper-parameter.

\noindent \textbf{Multiple Intent Detection Decoder}\, Follow \cite{qin-etal-2020-agif}, after obtaining $\hat{\boldsymbol{r}}^{u^I}$ (\textit{i.e.}, $\hat{\boldsymbol{r}}^{u})$, we can use it for \textit{multiple intent detection}:
\begin{equation}
\setlength{\abovedisplayskip}{3pt}
\setlength{\belowdisplayskip}{3pt}
\scalebox{.95}{$
\boldsymbol{p}^I=\sigma (\boldsymbol{W}_I(\operatorname{LeakyReLU}(\boldsymbol{W}_u\hat{\boldsymbol{r}}^{u^I}+\boldsymbol{b}_u))+\boldsymbol{b}_I)
$}
\end{equation}
where $\boldsymbol{p}^I$ denotes intent probability distribution, $\sigma$ denotes the $\operatorname{sigmoid}$ activation function, $\boldsymbol{W}_*$ and $\boldsymbol{b}_*$ are trainable matrix parameters. Then, we apply the same mechanism as \cite{chen-etal-2022-transformer} to obtain intent number $O^{{\rm IND}}$ of the input utterance:
\begin{equation}
\setlength{\abovedisplayskip}{3pt}
\setlength{\belowdisplayskip}{3pt}
\scalebox{.95}{$
O^{{\rm IND}} = \operatorname{argmax} (\mathcal{F}(\boldsymbol{W}_{ind}\boldsymbol{h}_{\texttt{CLS}}+\boldsymbol{b}_{ind}))
$}
\end{equation}
where $\mathcal{F}$ denotes the $\operatorname{softmax}$ activation function, $\boldsymbol{W}_{ind}$ is a trainable matrix parameter.
% To get final intent labels, we apply the same mechanism as to get intent number $O^{IND}$ of the utterance:
After that, we choose the top $O^{{\rm IND}}$ in $\boldsymbol{p}^I$ as the final intent result $\boldsymbol{O}^I=(o_{1}^{I},...,o_{O^{{\rm IND}}}^{I})$.
\vspace{-0.75em}
\subsection{Dynamic Graph Interaction Layer}
\label{sec:2.4}
\vspace{-0.75em}

\noindent \textbf{Graph Construction}\, Mathematically, our graph can be denoted as $\mathcal{G} = (\mathcal{V},\mathcal{E} ) $ where vertices refer to intents and slots, edges refer to correlations between them.

% $G^{[S,1]}={\textstyle \sum_{i=1}^{n+2}} \hat{r}^{u^S}_i=\{\hat{h}_1,...,\hat{h}_{n+2}\}$
% To model the interaction between intent and slot token, w
\textbf{Vertices}\, We have $n + m$ number of nodes in the interactive graph where $n$ is the utterance length and $m$ is the number of intents $O^{{\rm IND}}$ in \S\ref{sec:2.3}. The input of slot token feature is $\boldsymbol{G}^{[S,1]}={\textstyle \sum_{i=1}^{n}}\hat{\boldsymbol{h}}_i$  while the input of  intent feature is $\boldsymbol{G}^{[I,1]}= \{\phi ^{emb}(o_{1}^{I}),...,\phi ^{emb}(o_{O^{{\rm IND}}}^{I}) \}$ where $\phi ^{emb}$ is the embedding mapping function to map $o_{*}^{I}$ to its embedding $\boldsymbol{r}_{*}^{\varphi}$. The first layer states vector for two kind of nodes is $\boldsymbol{G}^1=\{\boldsymbol{G}^{[I,1]}, \boldsymbol{G}^{[S,1]}\}=\{\phi ^{emb}(o_{1}^{I}),...,\phi ^{emb}(o_{O^{{\rm IND}}}^{I}),\hat{\boldsymbol{h}}_1,...,\hat{\boldsymbol{h}}_{n} \}$.

\textbf{Edges}\, There are three types of connections in this graph:\\
\EmbracOff
\textsl{(a)}
\EmbracOn \texttt{intent}-\texttt{intent} connection: We connect all intent nodes to each other to model the relationship between each intent, since all of them appear in the same utterance. \\
\EmbracOff
\textsl{(b)}
\EmbracOn \texttt{slot}-\texttt{slot} connection: We connect each slot node to other slots with a window size to further model slot dependency and incorporate bidirectional contextual information.\\
\EmbracOff
\textsl{(c)}
\EmbracOn \texttt{intent}-\texttt{slot} connection: We adopt a scaled dot-product attention mechanism \cite{vaswani2017attention} for computing relevance between intent and token as follows:
\begin{equation}
\setlength{\abovedisplayskip}{3pt}
\setlength{\belowdisplayskip}{3pt}
\scalebox{1.25}{$
\delta _{ij}=\frac{\exp(\hat{\boldsymbol{h}}_i\phi ^{emb}(o_{j}^{I})^\top/\sqrt{d})}{ {\sum_{k=1}^{n}\exp(\hat{\boldsymbol{h}}_k\phi ^{emb}(o_{j}^{I})^\top/\sqrt{d})} } 
$}
\end{equation}
where $d$ is the dimension of hidden states, $\delta _{ij}$ is the relevance score between $i$-th token and $j$-th intent. We innovatively employ a hyper-parameter $\delta$ to measure intent and token relevance. If $\delta_{ij} > \delta$, it indicates that the token plays a significant role in determining the intent. In this case, this token is directly connected to the relevant intent (\textit{cf.} Fig.\ref{fig:network} \textit{right}).

\noindent \textbf{Graph Network}\, We use Graph Attention Network (GAT) \cite{velivckovic2017graph} to model intent-slot interaction. Specifically, for a given graph with $n$ nodes, GAT take the initial node features $\tilde{\boldsymbol{H}}=\{\tilde{\boldsymbol{h}_1},...\tilde{\boldsymbol{h}_n}\} $ as input to produce more abstract representation $\tilde{\boldsymbol{H}'}=\{\tilde{\boldsymbol{h}_1'},...\tilde{\boldsymbol{h}_n'}\} $ as its output. Within the graph, the aggregation at $l$-th layer can be defined as:
\begin{equation}
\setlength{\abovedisplayskip}{3pt}
\setlength{\belowdisplayskip}{3pt}
\scalebox{.95}{$
\boldsymbol{g}_{i}^{S,l+1} = \sigma (\textstyle \sum_{j\in\boldsymbol{\mathcal{G}}^S }^{}\alpha _{ij}\boldsymbol{W}_g\boldsymbol{g}_{j}^{[S,l]}+ \textstyle \sum_{j\in\boldsymbol{\mathcal{G}}^I }^{}\alpha _{ij}\boldsymbol{W}_g\boldsymbol{g}_{j}^{[I,l]}  )
$}
\end{equation}
where $\alpha_{ij}$ is the attention coefficient, $\boldsymbol{\mathcal{G}}^S$ and $\boldsymbol{\mathcal{G}}^I$ are vertices sets which denotes connected slots and intents, respectively.

% As shown in Fig.\ref{fig:network} (b)
\noindent \textbf{Slot Filling Decoder}\, After $L$ layers' propagation, we obtain the final slot representation $\boldsymbol{G}^{[S,L+1]}$ for \textit{slot prediction}:
\begin{equation}
\setlength{\abovedisplayskip}{3pt}
\setlength{\belowdisplayskip}{3pt}
\scalebox{.95}{$
O^S_t=\operatorname{argmax}(\mathcal{F}(\boldsymbol{W}_S\boldsymbol{g}_{t}^{[S,L+1]}+\boldsymbol{b}_S) )
$}
\end{equation}
where $\boldsymbol{W}_S$ is a trainable parameter and $O^S_t$ is the predicted slot of the $t$-th token in an utterance.
\vspace{-0.75em}
\subsection{Joint Training}
\vspace{-0.75em}
We adopt joint training to learn parameters. \textit{Multiple intent} and its \textit{number detection} are trained with binary cross-entropy while \textit{slot filling} is trained with cross-entropy. The final joint objective is formulated as:
\begin{equation}
\setlength{\abovedisplayskip}{3pt}
\setlength{\belowdisplayskip}{3pt}
\scalebox{.95}{$
\mathcal{L}=\alpha (\mathcal{L}_{{\rm ID}} + \gamma  \mathcal{L}_{{\rm RE}}^{I}) + \beta (\mathcal{L}_{{\rm SF}} + \gamma  \mathcal{L}_{{\rm RE}}^{S})+(1-\alpha )\mathcal{L}_{{\rm IND}}
$}
\end{equation}
where $\alpha$, $\beta$ and $\gamma$ are trade-off hyper-parameters.

\section{Experiments}
\label{sec:exp}
\vspace{-0.75em}
\subsection{Datasets}
\vspace{-0.75em}
We conduct experiments on two public multi-intent SLU datasets\footnote{\label{v2}\url{https://github.com/LooperXX/AGIF}}. \textbf{MixATIS} \cite{qin2021gl} is constructed from ATIS \cite{hemphill1990atis}, containing 13,162/756/828 utterances for train/validation/test. \textbf{MixSNIPS} \cite{qin2021gl,coucke2018snips} contains 39,776/2,198/2,199 utterances for train/validation/test. In addition, both of datasets are the cleaned version.

% \begin{table}[htbp]
% \vspace{-0.5em}
%   \centering
%   \fontsize{8.75}{10}\selectfont
%     \begin{tabular}{l|c|c|c}
%     \hline
%      \diagbox{\textbf{Dataset}}{\textbf{\# of Sentences}}{\textbf{\# of Intents}} & \textbf{1} & \textbf{2} & \textbf{3}    \\
%      \hline
%      MixATIS & 1365 & 9385 & 3996 \\
%      MixSNIPS & 9175 & 24998 & 10000 \\
%     \hline
%     \end{tabular}%
%   \caption{Summary of the amount of utterances in MixATIS and MixSNIPS with varying intent numbers.}
%   \label{tab:dataset1}%
% \vspace{-1.5em}
% \end{table}%

\vspace{-0.75em}
\subsection{Experimental Settings}
\vspace{-0.75em}
Considering the inference speed, we use English uncased \texttt{BERT-Base} model \cite{devlin2018bert} which consists of 12 layers, 12 heads and 768 hidden states. The batch size is 32 and the epoch is 50. Adam is used for optimization with learning rate of $2e$-$5$. The layer number of GAT is set to 2. For hyper-parameters of loss $\alpha$, $\beta$ and $\gamma$ are empirically set as 0.6: 1: 0.3 for MixATIS and 0.8: 1.2: 0.2 for MixSNIPS. 

% The model which works the best on the dev set will be chosen, and then we evaluate it on the test set.

% \begin{table}[htbp]
% \vspace{-1.0em}
%   \centering
%     \begin{tabular}{l|c}
%     \toprule
%      \textbf{Hyper-parameter} & \textbf{Search Range}  \\
%     \midrule
%     Learning Rate & \{1e-5, \textbf{2e-5}, 5e-5\} \\
%     Dropout Rate & \{0, 0.1, \textbf{0.2}, 0.3, 0.4\}  \\
%     $\alpha$ in Loss & \{0.6, \textbf{0.8}, 1\} \\
%     $\beta$ in Loss & \{1, \textbf{1.2}, 1.4\}\\
%     $\gamma$ in Loss & \{0.1, \textbf{0.3}, 0.5\} \\
%     \bottomrule
%     \end{tabular}%
%   \caption{Hyper-parameter search range of LDGF. \textbf{Bold} numbers indicate our choice of hyper-parameters.}
%   \label{tab:dataset}%
% \vspace{-1.0em}
% \end{table}%

\vspace{-0.75em}
\subsection{Baselines}
\vspace{-0.75em}
We compare our \texttt{DGIF} with both single-intent and multi-intent models. For single-intent SLU models to handle multi-intent utterances, multiple intents are concatenated with ‘\#’ into a single intent for a fair comparison. We also obtain pre-trained language models (PLMs) for comparison which are \texttt{Bert-base} and \texttt{Roberta-base}.
\renewcommand{\arraystretch}{1.2} %控制行高
\begin{table*}[!ht]
% \vspace{-0.5em}
\centering
\fontsize{8}{8.5}\selectfont
\begin{threeparttable}
\begin{tabular}{lcccccc}
\toprule
\multirow{3}{*}{Model}&
\multicolumn{3}{c}{ MixATIS}&\multicolumn{3}{c}{ MixSNIPS}\cr
\cmidrule(lr){2-4} \cmidrule(lr){5-7}
&Slot (F1)&Intent (Acc)&Overall (Acc)&Slot (F1) &Intent (Acc)&Overall (Acc)\cr
\midrule
Bi-Model$_{\it(2018)}$ \cite{wang2018bi}& 83.9& 70.3& 34.4& 90.7& 95.6& 63.4\cr
SF-ID$_{\it(2019)}$ \cite{haihong2019novel}& 87.4& 66.2& 34.9& 90.6& 95.0& 59.9\cr
Stack-Propagation$_{\it(2019)}$ \cite{qin2019stack} & 87.8& 72.1& 40.1& 94.2& 96.0& 72.9\cr
Joint Multiple ID-SF$_{\it(2019)}$ \cite{gangadharaiah2019joint}  & 84.6& 73.4& 36.1& 90.6& 95.1& 62.9\cr
AGIF$_{\it(2020)}$ \cite{qin-etal-2020-agif} & 86.7& 74.4& 40.8& 94.2& 95.1& 74.2\cr
GL-GIN$_{\it(2021)}$ \cite{qin2021gl} &\uline{88.3} & 76.3& 43.5&94.9 &95.6& 75.4\cr
SDJN$_{\it(2022)}$ \cite{chen2022joint} &88.2 & 77.1& 44.6&94.4 &96.5& 75.7\cr
\midrule
% \hdashline

Bert-baseline &86.3 & 74.5& 44.8&\uline{95.5} &95.6& 80.1\cr
Roberta-baseline &85.0 & \uline{78.3}& \uline{47.8}&\textbf{95.9} &\uline{97.5}& \uline{83.2}\cr
\midrule
% \hdashline
\textbf{DGIF} &\textbf{88.5}$^\dag$ &{\bf83.3}$^\dag$&{\bf50.7}$^\dag$&{\bf95.9}&{\bf97.8}$^\dag$& {\bf84.3}$^\dag$\cr
\bottomrule
\end{tabular}

\end{threeparttable}
\caption{Main results on two multi-intent datasets. The best performance is in \textbf{bold} and the second best performance is \uline{underlined}. $\dag$ indicate that the improvement of \texttt{DGIF} over all baselines is statistically significant with $p < 0.05$ under t-test.  }
\label{tab:main_results}
\vspace{-1.0em}
\end{table*}
\renewcommand{\arraystretch}{1.05} %控制行高
\begin{table}[!ht]
% \vspace{-1.0em}
\centering
\fontsize{8.75}{10}\selectfont
\begin{threeparttable}
\begin{tabular}{lccc}
\toprule
\multirow{3}{*}{Model}&
\multicolumn{3}{c}{ MixATIS}\cr
\cmidrule(lr){2-4} 
&Slot&Intent&Overall\cr
&(F1)&(Acc)&(Acc)\cr
\midrule
\textbf{DGIF} & \textbf{88.5}& \textbf{83.3}& \textbf{50.7}\cr
\midrule
\textit{w/o LAR} & 87.9 ($\downarrow$0.6) & 81.2 ($\downarrow$2.1) & 49.6 ($\downarrow$1.1)\cr
\textit{w/o LAR + LSI} & 87.1 ($\downarrow$1.4) & 77.5 ($\downarrow$5.8) & 47.3 ($\downarrow$3.4)\cr
\hdashline
\textit{w/o LAR + LSI + GIL} & 86.4 ($\downarrow$2.1) & 75.6 ($\downarrow$7.7) & 45.4 ($\downarrow$5.3)\cr
\bottomrule
\end{tabular}
\end{threeparttable}
\caption{Results of ablation test on the MixATIS dataset.}
\label{tab:ablation}
\vspace{-1.75em}
\end{table}Following \cite{chen2019bert}, we obtain the hidden state of the first special token (\texttt{[CLS]}) with $\operatorname{sigmoid}$ function for detecting multi-intent and use hidden states of utterance tokens for slot filling.

\vspace{-0.75em}
\subsection{Main Results}
\vspace{-0.75em}
% We evaluate the performance using three metrics: F1 score of slot filling (Slot F1), accuracy of intent detection (Intent Acc) and sentence-level overall accuracy (Overall Acc) as in \cite{goo2018slot,qin2021gl}. 
We evaluate performance using three metrics: F1 score of slot filling (Slot F1), accuracy of intent detection (Intent Acc) and sentence-level overall accuracy (Overall Acc) as in \cite{qin2021gl,goo2018slot}. “Overall Acc” considers prediction of an utterance is correct only when its intents and slots are all correctly predicted.

Table \ref{tab:main_results} shows experiment results of different models on the MixATIS and MixSNIPS datasets. From the results, we have the following observations: (1) For \textit{slot F1}, our method leads to slight imrovements (0.2\% and 1\%) compared to the best baseline \texttt{GL-GIN}, which validates that \texttt{DGIF} is more effective on slot filling. (2) Turning to \textit{intent accuracy}, \texttt{DGIF} exceeds \texttt{SDJN} by 6.2\% and 1.3\%, respectively. It proves that \texttt{DGIF} has a strong ability to identify intents. (3) Moreover, \texttt{DGIF} surpasses \texttt{SDJN} 6.1\% and 8.6\% on \textit{overall accuracy}, which confirms that \texttt{DGIF} is more powerful in understanding the implicit correlations between intents and slots. 

We attribute the above gains to our approach injecting label semantics into utterance to provide additional signals and enriched priors than \textit{one-hot} encoded labels. On the basis of semantic information injection, our multi-grain interactive graph helps the model capture the more \textbf{explicit} and \textbf{high-confidence} correspondence between intents and slots to alleviate error propagation, which brings significant improvements to multi-intent SLU.

In particular, we can find that although \texttt{Roberta-base} brings remarkable improvements via generating high-quality word representations than \texttt{Bert-base}, our  \texttt{Bert-base} \texttt{DGIF} significantly outperforms its counterparts. We intuitively suspect this is because our approach can capture and leverage the correlations based on label-semantic injection among the intent and slot labels. By this means, the correct predictions of the two tasks can be better aligned.

\vspace{-0.75em}
\subsection{Analysis}
\vspace{-0.75em}
\subsubsection{Ablation Test}
\vspace{-0.75em}
To verify effectiveness of each component in \texttt{DGIF}, we conduct a set of ablation experiments on the MixATIS dataset.

\noindent \textbf{Effectiveness of Adaptive Label-Semantic Injection Layer}\, \\
(1) We remove label-aware regularization, which is named as \textit{w/o LAR} in Table \ref{tab:ablation}. From the results, we can observe the absence of label-aware regularization leads to 0.6\% \textit{slot F1} and 2.1\% \textit{intent accuracy} drops. This indicates that the label-aware regularization encourages our network to model rich semantic dependencies among labels in each label space. \\
(2) We further remove label-semantic injection and utilizes the product of utterance/token representations and labels and keep other components unchanged. It is named as \textit{w/o LAR + LSI} in Table \ref{tab:ablation}. 
% \begin{figure}[!ht]
% \fontsize{8}{8.5}\selectfont
% \vspace{-1em}
% \begin{minipage}[a]{1\linewidth}
%   \centering
%   \centerline{\includegraphics[width=8.5cm]{f1.pdf}}
% \vspace{-0.2em}
%   \centerline{(a) The attention distribution of AGIF \cite{qin-etal-2020-agif}.}\smallskip
% \end{minipage}
% \vspace{-0.75em}
% \begin{minipage}[b]{1\linewidth}
%   \centering
%   \centerline{\includegraphics[width=8.5cm]{f1.pdf}}
% \vspace{-0.2em}
%     \centerline{(b) The attention distribution of our model.}\smallskip
% \end{minipage}
% \vspace{-1.5em}
% \caption{Attention heatmap in different approaches.}
% \label{fig:visualization}
% \vspace{-2.25em}
% \end{figure}
\begin{figure}[!ht]
% \fontsize{8}{8.5}\selectfont
% \vspace{-1em}
\begin{minipage}[a]{1.0\linewidth}
  \centering
  \centerline{\includegraphics[width=9cm]{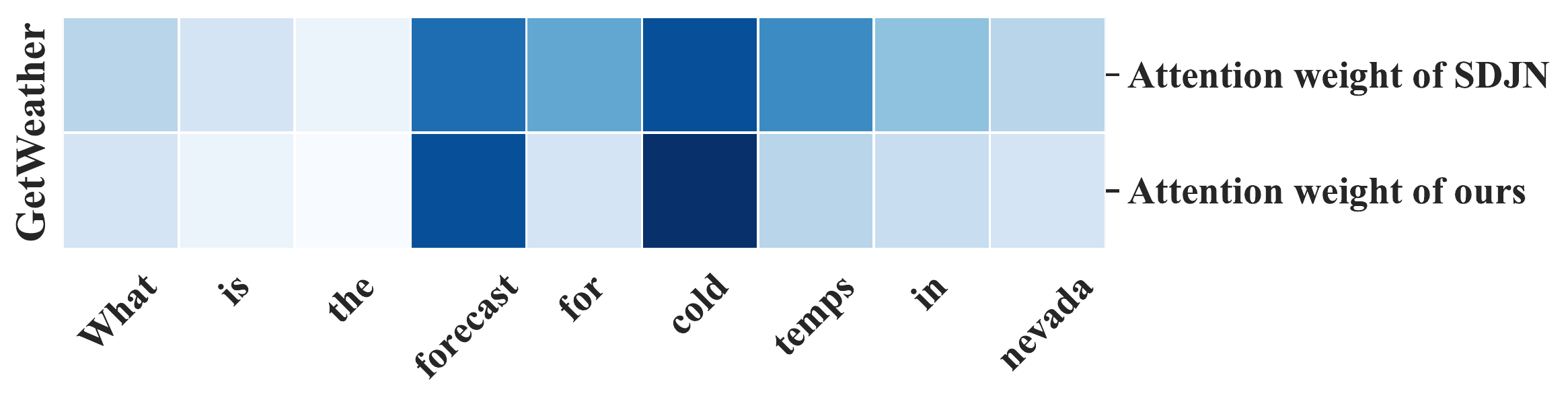}}
    \smallskip
\end{minipage}
\vspace{-1em}
\caption{Attention heatmap in different approaches.}
\label{fig:visualization}
\vspace{-1.5em}
\end{figure}We can obviously observe that the \textit{overall accuracy} drops by 3.4\%. This indicates that label-semantic injection can capture the correlation between utterances and explicit labels' semantics, which is beneficial for the semantic performance of multi-intent SLU system.

\noindent \textbf{Effectiveness of Dynamic Graph Interaction Layer}\, On the basis of the previous experiments, we replace the dynamic interactive layer with the vanilla attention mechanism, which is named as \textit{w/o LAR + LSI + GIL} in Table \ref{tab:ablation}. We can observe the performance drops in all metrics on the MixATIS dataset. We attribute it to the fact that our approach can automatically filter irrelevant intent information for each token.

\vspace{-0.75em}
\subsubsection{Visualization}
\vspace{-0.75em}
To better understand what the interactive graph layer has learned, we visualize the attention weight of it and \texttt{SDJN} \cite{chen2022joint} counterpart version for comparison, which is shown in Fig. \ref{fig:visualization}. We can clearly observe that our framework properly 
aggregates relevant intent “\texttt{GetWeather}” at slots “\texttt{forecast}” and “\texttt{cold}” respectively where the attention weights successfully focus on the correct slot. This justifies our \texttt{DGIF} has a better interaction ability compared to the prior approach. 
% \begin{figure}[!ht]
% \begin{minipage}[a]{1.0\linewidth}
%   \centering
%   \centerline{\includegraphics[width=8.5cm]{f1.pdf}}
%   \smallskip
% \end{minipage}
% \caption{Visualization.}
% \label{fig:visualization}
% \vspace{-1em}
% \end{figure}

\section{Conclusion}
\vspace{-0.75em}
In this paper, we propose regularized label semantics injection for joint \textit{multiple intent detection} and \textit{slot filling}. By considering label semantics, we devise a novel approach to construct a multi-grain graph for dynamic interaction. Experimental analyses on two public multi-intent datasets verify the effectiveness of our approach.

\section{Acknowledgements}
\vspace{-0.75em}
% We also thank all anonymous reviewers for their constructive comments.
This paper was partially supported by Shenzhen Science \& Technology Research Program (No:GXWD2020123116580\\7007-20200814115301001) and NSFC (No: 62176008). 
% This paper was partially supported by Shenzhen Science \& Technology Research Program (No: GXWD2020123116580\\7007-20200814115301001) and NSFC (No: 62176008).
% \vfill\pagebreak

% \section{REFERENCES}
% \label{sec:refs}

% List and number all bibliographical references at the end of the
% paper. The references can be numbered in alphabetic order or in
% order of appearance in the document. When referring to them in
% the text, type the corresponding reference number in square
% brackets as shown at the end of this sentence \cite{C2}. An
% additional final page (the fifth page, in most cases) is
% allowed, but must contain only references to the prior
% literature.

\normalem
\bibliographystyle{IEEEbib}
\bibliography{strings,refs}

\end{document}